\documentclass{article} % For LaTeX2e
\usepackage{nips13submit_e,times}
\usepackage[numbers]{natbib}
\usepackage{hyperref}
\usepackage{url}
\usepackage{graphicx}
\usepackage{amsmath}
\usepackage{amsfonts}
\usepackage{subcaption}
\usepackage[font=small]{caption}
\usepackage{authblk}

\title{Pylearn2: a machine learning research library}

\author[1]{Ian J. Goodfellow}
\author[1]{David Warde-Farley}
\author[1]{Pascal Lamblin}
\author[1]{Vincent Dumoulin}
\author[1]{Mehdi Mirza}
\author[1]{Razvan Pascanu}
\author[2]{James Bergstra}
\author[1]{Fr\'ed\'eric Bastien}
\author[1]{Yoshua Bengio}
\affil[1]{D\'{e}partement d'Informatique et de Recherche Op\'{e}rationelle\\
Universit\'{e} de Montr\'{e}al \texttt{\{goodfeli, wardefar, lamblinp, dumouliv, mirzamom, pascanur\}@iro.umontreal.ca, nouiz@nouiz.org, yoshua.bengio@umontreal.ca}}
\affil[2]{Center for Theoretical Neuroscience, University of Waterloo. \texttt{james.bergstra@uwaterloo.ca}}
\nipsfinalcopy % Uncomment for camera-ready version

\begin{document}

\maketitle

\begin{abstract}
Pylearn2 is a machine learning research library.
This does not just mean that it is a collection of machine
learning algorithms that share a common API; it means that
it has been designed for flexibility and extensibility in
order to facilitate research projects that involve new or
unusual use cases. In this paper we give a brief history
of the library, an overview of its basic philosophy, a
summary of the library's architecture, and a description of
how the Pylearn2 community functions socially.
\end{abstract}

\section{Introduction}

Pylearn2 is a machine learning research library developed by LISA
at Universit\'{e} de Montr\'{e}al. The goal of the library is to
facilitate machine learning research. This means that the library
has a focus on flexibility and extensibility, in order to make sure
that nearly any research idea is feasible to implement in the library.
The target user base is machine learning researchers.
Being ``user friendly'' for a research
user means that it should be easy to understand exactly what the code is
doing and configure it very precisely for any desired experiment.
Sometimes this may come at the cost of requiring the user to be an expert practitioner,
who must understand how the algorithm works in order to accomplish basic
data analysis tasks. This is
different from other notable machine learning libraries, such as scikit-learn~\citep{scikit-learn}
or the learning algorithms provided as part of OpenCV~\citep{opencv_library},
the STAIR Vision Library~\citep{gouldstair}, etc. Such machine learning libraries aim to provide good
performance to users who do not necessarily understand how the underlying algorithm
works. Pylearn2 has a different user base, and thus different design goals.

In this paper, we give a general sense of the library's design and how the community
functions. We begin with a brief history of the library.
Finally, we give an overview of the library's philosophy, the architecture of the library
itself, and the development workflow that the Pylearn2 community uses to improve the library.

\begin{table}[ht]
\begin{tabular}{c|c}
GitHub repository & \url{https://github.com/lisa-lab/pylearn2} \\
Documentation & \url{http://deeplearning.net/software/pylearn2} \\
User mailing list & \url{pylearn-users@googlegroups.com} \\
Developer mailing list & \url{pylearn-dev@googlegroups.com} \\
\end{tabular}
\caption{Other Pylearn2 resources}
\end{table}

\section{History}
Pylearn2 is LISA's third major effort to design a flexible machine learning
research library, the former two being PLearn and Pylearn. It is built on top of the lab's
mathematical expression compiler, Theano~\citep{bergstra+al:2010-scipy,Bastien-Theano-2012}.
In late 2010, a series of committees of LISA lab members met to plan how to fulfill LISA's
software development needs. These committees determined that no existing publicly available
machine library had design goals that would satisfy the requirements imposed by the kind of
research done at LISA\@. The committees decided to create Pylearn2, and drafted some basic
design ideas and the guiding philosophy of the library.
The first implementation work on Pylearn2 began as a class project in early 2011.
The library was used for research work mostly within LISA over the next two years.
In this time the structure of the library changed several times but eventually became stable.
The addition of Continuous Integration from Travis-CI~\citep{travis_ci}
with the development workflow from GitHub~\citep{github} helped to greatly improve the stability
of the library. GitHub provides a useful interface for reviewing code before it is merged, and
Travis-CI tells reviewers whether the code passes the tests.

In late 2011 Pylearn2 was used to win a transfer learning contest~\citep{Goodfeli-et-al-TPAMI-Deep-PrePrint-2013}.
After this, a handful of researchers outside LISA became interested in using it to reproduce 
the results from this challenge. However, the majority of Pylearn2 users were still LISA members.
Pylearn2 first gained a significant user base outside LISA in
the first half of 2013. This was in part due to the attention the library received after it was used
to set the state of the art on several computer vision benchmark tasks~\citep{Goodfellow-et-al-ICML2013},
and in part due to many Kagglers starting to use the library after the baseline solutions to some
Kaggle contests were provided in Pylearn2 format~\citep{Goodfeli-et-al-ICONIP-2013}.

Today, over 250 GitHub users watch the repository, nearly 200 subscribe to the mailing list,
and over 100 have made their own fork to work on new features. Over 30 GitHub users have
contributed to the library. 

\section{License and citation information}

Pylearn2 is released under the 3-claused BSD license, so it may be used for
commercial purposes. The license does not require anyone to cite Pylearn2, but
if you use Pylearn2 in published research work we encourage you to cite this
article.

\section{Philosophy}

Development of Pylearn2 is guided by several principles:

\begin{itemize}
\item {\bf Pylearn2 is a machine learning research library--its users are researchers.} This means the Pylearn2
framework should not impose many restrictions on what is possible to do with the library, and it is acceptable
to assume that the user has some technical sophistication and knowledge of machine learning.
\item {\bf Pylearn2 is built from re-usable parts, that can be used in many combinations or independently.} In
particular, no user should be forced to learn all parts of the library. If a user wants only to use a Pylearn2
\texttt{Model}, it should be possible to do so without learning about Pylearn2 \texttt{TrainingAlgorithm}s, 
\texttt{Cost}s, etc.
\item {\bf Pylearn2 avoids over-planning.}
Each feature is designed with an eye toward allowing more features to be developed in the future.
We do enough planning to ensure that our designs are modular and easy to extend. We generally do not
do much more planning than that.
This avoids paralysis and over-engineering.
\item {\bf Pylearn2 provides a domain-specific language that provides a compact way of specifying all hyperparameters
for an experiment.} Pylearn2 accomplishes this using YAML with a few extensions. A brief YAML file can instantiate
a complex experiment without exposing any implementation-specific detail. This makes it easier for researchers who
do not use Pylearn2 to read the specification of a Pylearn2 experiment and reproduce it using their own software.
\end{itemize}

\section{Library overview}

Pylearn2 consists of several components that can be combined to form complete learning
algorithms. Most components do not actually execute any numerical code--they just provide
symbolic expressions. This is possible because Pylearn2 is built on top of
Theano~\citep{bergstra+al:2010-scipy,Bastien-Theano-2012}.
Theano provides a language for describing expressions independent of how they are actually implemented,
so a single Pylearn2 class provides both CPU and GPU functionality. Another advantage of
using symbolic representations as the main arguments to Pylearn2 methods is that it is possible
to compute many functions of a symbolic expression that can not be computed from a numerical value alone.
For example, it is possible to compute the derivative of a Theano expression, while it is not possible
to compute the derivative of the process that generated a numerical value given only the value itself.
This means that many interfaces are simpler--few expressions need to be passed between objects, because
the recipient can create its own modifications of the expression it is passed, rather than needing an
interface for each modified value it requires.

\subsection{Core components}

The main way the Pylearn2 achieves flexibility and extensibility is decomposition
into reusable parts. The three key components used to implement most features are
the \texttt{Dataset}, \texttt{Model}, and \texttt{TrainingAlgorithm} classes.
A \texttt{Dataset} provides the data to be trained on. A \texttt{Model} stores
parameters and can generate Theano expressions that perform some useful task
given input data (e.g., estimate the probability density at a point in space,
infer a class label given input features, etc.).  A \texttt{TrainingAlgorithm}
adapts a \texttt{Model} to a particular \texttt{Dataset}. Generally each of these
objects is in turn modular (\texttt{Dataset}s have modular preprocessing, many \texttt{Model}
classes are organized into \texttt{Layer}s, \texttt{TrainingAlgorithms} can minimize
a modular \texttt{Cost} and can have their behavior modified by various modular callbacks
and a modular \texttt{TerminationCriterion}, etc.). This modularity means that if a researcher
has an innovative idea to test out, and that idea only affects one component, the researcher
can simply replace or subclass the component in question. The vast majority of the learning
system can still be used with the new idea.

This modularity is in contrast to most other machine learning libraries, where the \texttt{Model}
generally does most of the work. A scikit-learn model is generally accompanied by a \texttt{fit} method
that is a complete training algorithm and that can't be applied to any other model.
Some libraries are more modular but don't entirely divide the labor
between models and training algorithms as sharply as Pylearn2 does. For example, in Torch~\citep{Torch-2011}
or DistBelief~\citep{Dean-et-al-NIPS2012} the \texttt{Model}s are modular, but to train a layer, the layer needs to
implement at the very least a backpropogation method for computing derivatives. 
In Pylearn2, the \texttt{Model} is only
responsible for creating symbolic expressions, which the \texttt{TrainingAlgorithm} may or may not
symbolically differentiate at a later time. (Individual Theano ops must still implement a \texttt{grad}
method, but a comparatively small number of basic ops can be used to implement the comparatively large
number of more complex models that appear in most machine learning libraries).

However, another aspect of Pylearn2's design philosophy is that no user should be forced to learn the
entire framework. It's possible to just implement a \texttt{train\_batch} method and have the \texttt{DefaultTrainingAlgorithm}
do nothing but serve the \texttt{Model} batches of data from the \texttt{Dataset}, or to ignore \texttt{TrainingAlgorithm}s altogether and just pass a \texttt{Dataset} to the \texttt{Model}'s \texttt{train\_all} method.

To facilitate code reuse, whenever possible, individual components that are shipped with the library aim to be as modular
and orthogonal as possible, relying on other existing components -- e.g., most \texttt{Model}s that become part of the
library will defer their learning functionality to an existing \texttt{TrainingAlgorithm} and/or \texttt{Cost}
unless sufficiently specialized as to be infeasible.

\subsection{The \texttt{Dataset} class}

\texttt{Dataset}s are essentially interfaces between sources of data in arbitrary format and the in-memory
array formats that Pylearn2 expects. All Pylearn2 \texttt{Dataset}s provide the same interface but can be
implemented to use any back-end format. Currently, all Pylearn2 \texttt{Dataset}s just read data from disk,
but in principle a \texttt{Dataset} could access live streaming data from the network or a peripheral device
like a webcam.

\texttt{Dataset}s allow the data to be presented in many formats, regardless of how it is stored. For example,
a minibatch of 64 different $32 \times 32$ pixel RGB images could be presented as a $64 \times 3072$ element
design matrix, or it could be presented as a $64 \times 32 \times 32 \times 3$ tensor. The choice of which
attribute to put on which axis can change to support different pieces of software too (for example, Theano
convolution prefers batch size $\times$ channels $\times$ rows $\times$ columns, while cuda-convnet~\citep{Krizhevsky-2012}
(which is wrapped in Pylearn2)
prefers channels $\times$ rows $\times$ columns $\times$ batch size). The data can also be presented in many
different orders, allowing iteration in sequential or different types of random order. \texttt{Dataset}s can
implement as many or as few iteration schemes as the implementor wants to--depending on the back end of the
data, not all iteration schemes are efficient or even possible (for example, if the iterator needs to read
from a network drive, random iteration may be very slow, and if the iterator reads live video from a webcam, there is
no way for it to visit the future).

Most \texttt{Dataset}s used in the deep learning community can be represented as a design matrix stored in
dense matrix format. For these datasets, implementing an appropriate \texttt{Dataset} object is very easy.
The implementer only needs to subclass the \texttt{DenseDesignMatrix} class and implement a constructor that
loads the desired data. If the data is already stored in NumPy~\citep{numpy-2007} or pickle format, it is not even
necessary to implement any new Python code to use the dataset.

Some datasets can be described as dense design matrices but are too big to fit into memory. Pylearn2 supports
this use case via the \texttt{DenseDesignMatrixPyTables}. To use this class, the data must be stored in HDF5
format on disk. 

Most \texttt{Dataset}s also support some kind of preprocessing that can modify the data after it has been loaded.

\subsubsection{Implemented \texttt{Dataset}s}

Pylearn2 currently contains wrappers for several datasets. These include the datasets
used for DARPA's unsupervised and transfer learning challenge~\citep{Guyon-UTLC-ijcnn2011}, the
dataset used for the NIPS 2011 workshops challenge in representation
learning~\citep{NipsWorkshop11Hierarchical},
the CIFAR-10 and CIFAR-100 datasets~\citep{KrizhevskyHinton2009}, the MNIST dataset~\citep{LeCun+98},
some of the MNIST variations datasets~\citep{LarochelleH2007}, the NORB dataset~\citep{lecun-04}, the
Street View House Numbers dataset~\citep{Netzer-wkshp-2011}, the Toronto Faces Database~\citep{Susskind2010},
some of the UCI repository datasets~\citep{Fisher-1936,Diaconis:1983:CMS},
and a dataset of 3D animations of fish~\citep{Franzius2008}. Additionally, there are many kinds
of preprocessing, such as PCA~\citep{Pearson-1901}, ZCA~\citep{BellSejnowski-97}, various kinds of local contrast
normalization~\citep{sermanet-icpr-12}, as well as helper functions to set up the entire preprocessing
pipeline from some well-known successful and documented systems~\citep{Coates2011,SalHinton09small}.

\subsection{The \texttt{Model} class}

A \texttt{Model} is any object that stores parameters (for the purpose of Pylearn2, a ``non-parametric''
model is just one with a variable number of parameters). The basic \texttt{Model} class has very few interface
elements. Subclasses of the \texttt{Model} class define richer interfaces. These interfaces define different
quantities that the \texttt{Model} can compute. For example, the \texttt{MLP} class provides an \texttt{fprop}
method that provides a symbolic expression for forward propagation in a multilayer perceptron. If the final
layer of the MLP is a softmax layer representing distributions over classes $y$, and the \texttt{fprop} method
is passed a Theano variable representing inputs $x$, the output will be a Theano variable representing
$p(y \mid x)$. 

The \texttt{Model} class is not required to know how to train itself, though many models do. 

\subsubsection{Linear operators, spaces, and convolution}

Linear operations are key parts of many machine learning models. The distinction between many important classes
of machine learning models is often nothing more than what specific structure of linear transformation they use.
For example, both MLPs and convolutional networks apply linear operators followed by a nonlinearity to transform
inputs into outputs. In the MLP, the linear operation is multiplication by a dense matrix. In a convolutional
network, the linear operation is discrete convolution with finite support. This operation can be viewed as matrix
multiplication by a sparse matrix with several elements of the matrix constrained to be equal to each other. The
point is that both use a linear transformation. Pylearn2's \texttt{LinearTransform} class provides a generic representation of linear
operators. Pylearn2 functionality written using this class can thus be written once and then extended to do dense
matrix multiplication, convolution, tiled convolution, local connections, etc.\ simply by providing different
implementations of the linear operator. This idea grew out of James Bergstra's Theano-linear module which has since
been incorporated into Pylearn2.

Different linear operator implementations require their inputs to be formatted in different ways. For example,
convolution applied to an image requires a format that indicates the 2D position of each element of the input,
while dense matrix multiplication just requires a linearized vector representaiton of the image. In Pylearn2,
classes called \texttt{Space}s represent these different views of the same underlying data. Dense matrix multiplication
acts on data that lives in a \texttt{VectorSpace}, while 2D convolution acts on data that lives in a
\texttt{Conv2DSpace}. \texttt{Space}s generally know how to convert between each other, when possible. For example, an
image in a \texttt{Conv2DSpace} can be flattened into a vector in a \texttt{VectorSpace}.

Several linear operators (and related convolutional network operations like spatial max pooling) in Pylearn2
are implemented as wrappers that add Theano semantics on top of the extremely fast cuda-convnet
library~\citep{Krizhevsky-2012}, making Pylearn2 a very practical library to use for
convolutional network research.

\subsubsection{Implemented models}

Because the philosophy that Pylearn2 developers should write features when they are needed, and because
most Pylearn2 developers so far have been deep learning researchers, Pylearn2 mostly contains deep learning
models or models that are used as building blocks for deep architectures. This includes
autoencoders~\citep{Bourlard88},
RBMs~\citep{Smolensky86} including RBMs with Gaussian visible units~\citep{Welling05},
DBMs~\citep{SalHinton09small}, MLPs~\citep{Rumelhart86c},
convolutional networks~\citep{lecun-bengio-95a}, and local coordinate coding~\citep{Yu+al-NIPS09}.
However, Pylearn2 is not restricted
to deep learning functionality. We encourage submissions of other machine learning models. Often, LISA researchers
work on problems whose scale exceeds that of the typical machine learning library user, so we occasionally implement
features for simpler algorithms, such as SVMs~\citep{Cortes95} with reduced memory consumption or
k-means~\citep{Steinhaus57}
with fast multicore training.

Pylearn2 has implementations of several models that were developed at LISA, including denoising auto-encoders (DAEs)~\citep{VincentPLarochelleH2008},
contractive auto-encoders (CAEs)~\citep{Rifai+al-2011} including higher-order CAEs~\citep{Salah+al-2011}, spike-and-slab RBMs (ssRBMs)~\citep{courville+bergstra+bengio:2011aistats} including ssRBMs with
pooling~\citep{Courville+al-2011}, reconstruction sampling autoencoders~\citep{Dauphin+al-2011},
and deep sparse rectifier nets~\citep{Glorot+al-AI-2011}.
Pylearn2 also contains models that were developed not just at LISA but originally developed using Pylearn2,
such as spike-and-slab sparse coding with parallel variational inference~\citep{Goodfeli-et-al-TPAMI-Deep-PrePrint-2013}
and maxout units for neural nets~\citep{Goodfellow-et-al-ICML2013}.

\subsection{The \texttt{TrainingAlgorithm} class}

The role of the \texttt{TrainingAlgorithm} class is to adjust the parameters stored in a
\texttt{Model} in order to adapt the model to a given \texttt{Dataset}. The \texttt{TrainingAlgorithm}
is also responsible for a few less important tasks, such as setting up the \texttt{Monitor} to record
various values throughout training (to make learning curves, essentially). The \texttt{TrainingAlgorithm}
is one of the very few Pylearn2 classes that actually performs numerical computation. It gathers Theano
expressions assembled by the \texttt{Model} and other classes, synthesizes them into expressions for learning
rules, compiles the learning rules into Theano functions that accomplish the learning, and executes the
Theano expressions. In fact, Pylearn2 \texttt{TrainingAlgorithms} are not even required to use Theano at all.
Some use a mixture of Theano and generic Python code; for example to perform line searches with the control
logic done with basic Python loops and branching but the numerical computation done by Theano.

Most \texttt{TrainingAlgorithms} support constrained optimization by asking the \texttt{Model} to project the
result of each learning update back into an allowed region. Many Pylearn2 models impose non-negativity
constraints on parameters that represent, for example, the conditional variance of some random variable, and
most Pylearn2 neural network layers support max norm constraints on the weights~\citep{Srebro05}.

\subsubsection{The \texttt{Cost} class}

Many training algorithms can be expressed as procedures for iteratively minimizing a cost
function. This provides another opportunity for sharing code between algorithms. The \texttt{Cost}
class represents a cost function independent of the algorithm used to minimize it, and each
\texttt{TrainingAlgorithm} is free to use this representation or not, depending on what is most
appropriate. A \texttt{Cost} is essentially just a class for generating a Theano expression
describing the cost function, but it has a few extra pieces of functionality. For example, it can add
monitoring channels that are relevant to the cost being minimized (for example, one popular cost is
the negative log likelihood of class labels under a softmax model--this cost automatically adds a 
monitoring channel that tracks the misclassification rate of the classifier). One extremely important
aspect of the \texttt{Cost} is that it has a \texttt{get\_gradients} method. Unlike Theano's \texttt{grad}
method, this method is not guaranteed to return accurate gradients. This allows many algorithms that use
approximate gradients to be implemented using the same machinery as algorithms that use the exact gradient.
For example, the persistent contrastive divergence~\citep{Younes1999,Tieleman08} algorithm minimizes an
intractable cost with
intractable gradients--the log likelihood of a Boltzmann machine. The Pylearn2 \texttt{Cost} for persistent contrastive
divergence returns \texttt{None} for the value of the cost function itself, to express that the cost function
can't be computed. However, the standard \texttt{SGD} class is still able to perform stochastic gradient descent
on the cost, because the \texttt{get\_gradients} method for the cost returns a sampling-based approximation
to the gradient. No special optimization class is needed to handle this seemingly exotic case.

Other costs implemented in the library include dropout~\cite{Hinton-et-al-arxiv2012}, contrastive divergence~\citep{Hinton-PoE-2000},
noise contrastive estimation~\citep{Gutmann+Hyvarinen-2010}, score matching~\citep{Hyvarinen-2005}, 
denoising score matching~\citep{Vincent-NC-2011}, softmax log likelihood for classification with MLPs,
and Gaussian log likelihood for regression with MLPs. Many additional simpler costs can
be added together with the \texttt{SumOfCosts} class to combine these primary costs with secondary costs
to add regularization, such as weight decay or sparsity regularization.

\subsubsection{Implemented \texttt{TrainingAlgorithm}s}

Currently, Pylearn2 contains three main \texttt{TrainingAlgorithm} classes. The \texttt{DefaultTrainingAlgorithm}
does nothing but serve minibatches of data to the \texttt{Model}'s default minibatch learning rule. The
\texttt{SGD} class does stochastic gradient descent on a \texttt{Cost}. This class supports extensions
including Polyak averaging~\citep{Polyak+Juditsky-1992}, momentum~\citep{Rumelhart86c}, and
early stopping.
The \texttt{BGD} class does
batch gradient descent (in practice, large minibatches) aka the method of steepest descent~\citep{debye1954collected}.
The \texttt{BGD} class is able to accumulate contributions to the gradient from several minibatches before making an
update, thereby enabling it to use batches that are too large to fit in memory. Optional flags enable the \texttt{BGD}
class
to implement other similar algorithms such as nonlinear conjugate gradient descent~\citep{polak1969note}.

\section{Development workflow and user community}

Pylearn2 has many kinds of users and developers. One need not be a Pylearn2 developer to do research
with Pylearn2. Pylearn2 is a valuable research tool even for people who do not need to develop any new
algorithms. The wide array of reference implementations available in Pylearn2 make it useful for
studying how existing algorithms behave under various conditions, or for obtaining baseline results
on new tasks.

Researchers who wish to implement new algorithms with Pylearn2 do not necessarily need to become
Pylearn2 developers either. It's common to develop experimental features privately in an offline
repository. It's also perfectly fine to share Pylearn2 classes as part of a 3rd party repository
rather than having them merged to the main Pylearn2 repository.

For those who do wish to contribute to Pylearn2, thank you! The process is designed to make sure
the library is as stable as possible. Developers should first write to \texttt{pylearn-dev@googlegroups.com}
to plan how to implement their feature. If the feature requires a change to existing APIs, it's important
to follow the best practices
guide\footnote{\url{http://deeplearning.net/software/pylearn2/api_change.html}}.
Once a plan is in place, developers should write the feature
in their own fork of Pylearn2 on GitHub, then submit a pull request to the main repository.
Our automated test suite will run on the pull request and indicate whether it is safe to merge. Pylearn2
developers will also review the pull request. When both the automatic tests and the reviewers are satisfied,
one of us will merge the pull request. Be sure to write to \texttt{pylearn-dev} to find a reviewer for your
pull request.

All kinds of pull requests are welcome--new features (provided that they have tests), config files for
important results, bug fixes, and tests for existing features.

\section{Conclusion}

This article has described the Pylearn2 library, including its history, design philosophy and goals, basic architecture, and developer workflow. We hope you find Pylearn2 useful in your research
and welcome your potential contributions to it.

%\subsubsection*{Acknowledgments}

%Use unnumbered third level headings for the acknowledgments. All
%acknowledgments go at the end of the paper. Do not include 
%acknowledgments in the anonymized submission, only in the 
% final paper. 

\bibliography{strings,strings-shorter,ml,aigaion-shorter,aigaion}

\begin{thebibliography}{}

\bibitem[git(2013)git]{github}
 (2013).
\newblock Git{H}ub.
\newblock \url{http://github.com}.

\bibitem[tra(2013)tra]{travis_ci}
 (2013).
\newblock Travis {C}{I}.
\newblock \url{http://travis-ci.org}.

\bibitem[Bastien {\em et~al.}(2012)Bastien, Lamblin, Pascanu, Bergstra,
  Goodfellow, Bergeron, Bouchard, and Bengio]{Bastien-Theano-2012}
Bastien, F., Lamblin, P., Pascanu, R., Bergstra, J., Goodfellow, I., Bergeron,
  A., Bouchard, N., and Bengio, Y. (2012).
\newblock Theano: new features and speed improvements.
\newblock Deep Learning and Unsupervised Feature Learning NIPS 2012 Workshop.

\bibitem[Bell and Sejnowski(1997)Bell and Sejnowski]{BellSejnowski-97}
Bell, A. and Sejnowski, T.~J. (1997).
\newblock The independent components of natural scenes are edge filters.
\newblock {\em Vision Research\/}, {\bf 37}, 3327--3338.

\bibitem[Bergstra {\em et~al.}(2010)Bergstra, Breuleux, Bastien, Lamblin,
  Pascanu, Desjardins, Turian, Warde-Farley, and
  Bengio]{bergstra+al:2010-scipy}
Bergstra, J., Breuleux, O., Bastien, F., Lamblin, P., Pascanu, R., Desjardins,
  G., Turian, J., Warde-Farley, D., and Bengio, Y. (2010).
\newblock Theano: a {CPU} and {GPU} math expression compiler.
\newblock In {\em Proceedings of the Python for Scientific Computing Conference
  ({SciPy})\/}.
\newblock Oral Presentation.

\bibitem[Bourlard and Kamp(1988)Bourlard and Kamp]{Bourlard88}
Bourlard, H. and Kamp, Y. (1988).
\newblock Auto-association by multilayer perceptrons and singular value
  decomposition.
\newblock {\em Biological Cybernetics\/}, {\bf 59}, 291--294.

\bibitem[Bradski(2000)Bradski]{opencv_library}
Bradski, G. (2000).
\newblock {The OpenCV Library}.
\newblock {\em Dr. Dobb's Journal of Software Tools\/}.

\bibitem[Coates {\em et~al.}(2011)Coates, Lee, and Ng]{Coates2011}
Coates, A., Lee, H., and Ng, A.~Y. (2011).
\newblock An analysis of single-layer networks in unsupervised feature
  learning.
\newblock In {\em Proceedings of the Thirteenth International Conference on
  Artificial Intelligence and Statistics (AISTATS 2011)\/}.

\bibitem[Collobert {\em et~al.}(2011)Collobert, Kavukcuoglu, and
  Farabet]{Torch-2011}
Collobert, R., Kavukcuoglu, K., and Farabet, C. (2011).
\newblock Torch7: A matlab-like environment for machine learning.
\newblock In {\em Big{L}earn, {NIPS} {W}orkshop\/}.

\bibitem[Cortes and Vapnik(1995)Cortes and Vapnik]{Cortes95}
Cortes, C. and Vapnik, V. (1995).
\newblock Support vector networks.
\newblock {\em Machine Learning\/}, {\bf 20}, 273--297.

\bibitem[Courville {\em et~al.}(2011a)Courville, Bergstra, and
  Bengio]{courville+bergstra+bengio:2011aistats}
Courville, A., Bergstra, J., and Bengio, Y. (2011a).
\newblock A spike and slab restricted {B}oltzmann machine.
\newblock In G.~Gordon, D.~Dunson, and M.~Dud{\`{\i}}k, editors, {\em
  Proceedings of the Fourteenth International Conference on Artificial
  Intelligence and Statistics\/}, volume~15 of {\em JMLR W{\&}CP\/}.
\newblock Recipient of People's Choice Award.

\bibitem[Courville {\em et~al.}(2011b)Courville, Bergstra, and
  Bengio]{Courville+al-2011}
Courville, A., Bergstra, J., and Bengio, Y. (2011b).
\newblock Unsupervised models of images by spike-and-slab {RBM}s.
\newblock In {\em Proceedings of theTwenty-eight International Conference on
  Machine Learning (ICML'11)\/}.

\bibitem[Dauphin {\em et~al.}(2011)Dauphin, Glorot, and
  Bengio]{Dauphin+al-2011}
Dauphin, Y., Glorot, X., and Bengio, Y. (2011).
\newblock Large-scale learning of embeddings with reconstruction sampling.
\newblock In {\em Proceedings of theTwenty-eight International Conference on
  Machine Learning (ICML'11)\/}.

\bibitem[Dean {\em et~al.}(2012)Dean, Corrado, Monga, Chen, Devin, Le, Mao,
  Ranzato, Senior, Tucker, Yang, and Ng]{Dean-et-al-NIPS2012}
Dean, J., Corrado, G., Monga, R., Chen, K., Devin, M., Le, Q., Mao, M.,
  Ranzato, M., Senior, A., Tucker, P., Yang, K., and Ng, A.~Y. (2012).
\newblock Large scale distributed deep networks.
\newblock In {\em NIPS'2012\/}.

\bibitem[Debye(1954)Debye]{debye1954collected}
Debye, P. (1954).
\newblock {\em The collected papers of Peter J.W. Debye\/}.
\newblock Ox Bow Press.

\bibitem[Diaconis and Efron(1983)Diaconis and Efron]{Diaconis:1983:CMS}
Diaconis, P. and Efron, B. (1983).
\newblock Computer-intensive methods in statistics.
\newblock {\bf 248}(5), 116--126, 128, 130.

\bibitem[Fisher(1936)Fisher]{Fisher-1936}
Fisher, R.~A. (1936).
\newblock The use of multiple measurements in taxonomic problems.
\newblock {\em Annals of Eugenics\/}, {\bf 7}, 179--188.

\bibitem[Franzius {\em et~al.}(2008)Franzius, Wilbert, and
  Wiskott]{Franzius2008}
Franzius, M., Wilbert, N., and Wiskott, L. (2008).
\newblock Invariant object recognition with slow feature analysis.
\newblock In {\em Proceedings of the 18th international conference on
  Artificial Neural Networks, Part I\/}, ICANN '08, pages 961--970, Berlin,
  Heidelberg. Springer-Verlag.

\bibitem[Glorot {\em et~al.}(2011)Glorot, Bordes, and
  Bengio]{Glorot+al-AI-2011}
Glorot, X., Bordes, A., and Bengio, Y. (2011).
\newblock Deep sparse rectifier neural networks.
\newblock In {\em JMLR W\&CP: Proceedings of the Fourteenth International
  Conference on Artificial Intelligence and Statistics (AISTATS 2011)\/}.

\bibitem[Goodfellow {\em et~al.}(2013a)Goodfellow, Erhan, Carrier, Courville,
  Mirza, Hamner, Cukierski, Tang, Thaler, Lee, Zhou, Ramaiah, Feng, Li, Wang,
  Athanasakis, Shawe-Taylor, Milakov, Park, Ionescu, Popescu, Grozea, Bergstra,
  Xie, Romaszko, Xu, Chuang, and Bengio]{Goodfeli-et-al-ICONIP-2013}
Goodfellow, I., Erhan, D., Carrier, P.-L., Courville, A., Mirza, M., Hamner,
  B., Cukierski, W., Tang, Y., Thaler, D., Lee, D.-H., Zhou, Y., Ramaiah, C.,
  Feng, F., Li, R., Wang, X., Athanasakis, D., Shawe-Taylor, J., Milakov, M.,
  Park, J., Ionescu, R., Popescu, M., Grozea, C., Bergstra, J., Xie, J.,
  Romaszko, L., Xu, B., Chuang, Z., and Bengio, Y. (2013a).
\newblock Challenges in representation learning: A report on three machine
  learning contests.
\newblock In {\em International Conference On Neural Information Processing\/}.

\bibitem[Goodfellow {\em et~al.}(2013b)Goodfellow, Warde-Farley, Mirza,
  Courville, and Bengio]{Goodfellow-et-al-ICML2013}
Goodfellow, I., Warde-Farley, D., Mirza, M., Courville, A., and Bengio, Y.
  (2013b).
\newblock Maxout networks.
\newblock In S.~Dasgupta and D.~McAllester, editors, {\em ICML'13\/}, page
  1319–1327.

\bibitem[Goodfellow {\em et~al.}(2013c)Goodfellow, Courville, and
  Bengio]{Goodfeli-et-al-TPAMI-Deep-PrePrint-2013}
Goodfellow, I., Courville, A., and Bengio, Y. (2013c).
\newblock Scaling up spike-and-slab models for unsupervised feature learning.
\newblock {\em IEEE Transactions on Pattern Analysis and Machine
  Intelligence\/}, {\bf 35}(8), 1902--1914.

\bibitem[Gould {\em et~al.}(2010)Gould, Russakovsky, Goodfellow, Baumstarck,
  Ng, and Koller]{gouldstair}
Gould, S., Russakovsky, O., Goodfellow, I., Baumstarck, P., Ng, A.~Y., and
  Koller, D. (2010).
\newblock The stair vision library.

\bibitem[Gutmann and Hyvarinen(2010)Gutmann and
  Hyvarinen]{Gutmann+Hyvarinen-2010}
Gutmann, M. and Hyvarinen, A. (2010).
\newblock Noise-contrastive estimation: A new estimation principle for
  unnormalized statistical models.
\newblock In {\em Proceedings of The Thirteenth International Conference on
  Artificial Intelligence and Statistics (AISTATS'10)\/}.

\bibitem[Guyon {\em et~al.}(2011)Guyon, Dror, Lemaire, Taylor, and
  Aha]{Guyon-UTLC-ijcnn2011}
Guyon, I., Dror, G., Lemaire, V., Taylor, G., and Aha, D.~W. (2011).
\newblock Unsupervised and transfer learning challenge.
\newblock In {\em Proc. Int. Joint Conf. on Neural Networks\/}.

\bibitem[Hinton(2000)Hinton]{Hinton-PoE-2000}
Hinton, G.~E. (2000).
\newblock Training products of experts by minimizing contrastive divergence.
\newblock Technical Report GCNU TR 2000-004, Gatsby Unit, University College
  London.

\bibitem[Hinton {\em et~al.}(2012)Hinton, Srivastava, Krizhevsky, Sutskever,
  and Salakhutdinov]{Hinton-et-al-arxiv2012}
Hinton, G.~E., Srivastava, N., Krizhevsky, A., Sutskever, I., and
  Salakhutdinov, R. (2012).
\newblock Improving neural networks by preventing co-adaptation of feature
  detectors.
\newblock Technical report, arXiv:1207.0580.

\bibitem[Hyv{\"{a}}rinen(2005)Hyv{\"{a}}rinen]{Hyvarinen-2005}
Hyv{\"{a}}rinen, A. (2005).
\newblock Estimation of non-normalized statistical models using score matching.
\newblock {\em Journal of Machine Learning Research\/}, {\bf 6}, 695--709.

\bibitem[Krizhevsky and Hinton(2009)Krizhevsky and
  Hinton]{KrizhevskyHinton2009}
Krizhevsky, A. and Hinton, G. (2009).
\newblock Learning multiple layers of features from tiny images.
\newblock Technical report, University of Toronto.

\bibitem[Krizhevsky {\em et~al.}(2012)Krizhevsky, Sutskever, and
  Hinton]{Krizhevsky-2012}
Krizhevsky, A., Sutskever, I., and Hinton, G. (2012).
\newblock {ImageNet} classification with deep convolutional neural networks.
\newblock In {\em Advances in Neural Information Processing Systems 25
  (NIPS'2012)\/}.

\bibitem[Larochelle {\em et~al.}(2007)Larochelle, Erhan, Courville, Bergstra,
  and Bengio]{LarochelleH2007}
Larochelle, H., Erhan, D., Courville, A., Bergstra, J., and Bengio, Y. (2007).
\newblock An empirical evaluation of deep architectures on problems with many
  factors of variation.
\newblock In {\em ICML'07\/}, pages 473--480. ACM.

\bibitem[Le {\em et~al.}(2011)Le, Ranzato, Salakhutdinov, Ng, and
  Tenenbaum]{NipsWorkshop11Hierarchical}
Le, Q.~V., Ranzato, M., Salakhutdinov, R., Ng, A., and Tenenbaum, J. (2011).
\newblock {\em NIPS Workshop on Challenges in Learning Hierarchical Models:
  Transfer Learning and Optimization\/}.
\newblock \url{https://sites.google.com/site/nips2011workshop}.

\bibitem[{LeCun} and Bengio(1995){LeCun} and Bengio]{lecun-bengio-95a}
{LeCun}, Y. and Bengio, Y. (1995).
\newblock Convolutional networks for images, speech, and time-series.
\newblock In M.~A. Arbib, editor, {\em The Handbook of Brain Theory and Neural
  Networks\/}, pages 255--257. MIT Press.

\bibitem[{LeCun} {\em et~al.}(1998){LeCun}, Bottou, Bengio, and
  Haffner]{LeCun+98}
{LeCun}, Y., Bottou, L., Bengio, Y., and Haffner, P. (1998).
\newblock Gradient-based learning applied to document recognition.
\newblock {\em Proceedings of the {IEEE}\/}, {\bf 86}(11), 2278--2324.

\bibitem[{LeCun} {\em et~al.}(2004){LeCun}, Huang, and Bottou]{lecun-04}
{LeCun}, Y., Huang, F.-J., and Bottou, L. (2004).
\newblock Learning methods for generic object recognition with invariance to
  pose and lighting.
\newblock In {\em Proceedings of the Computer Vision and Pattern Recognition
  Conference (CVPR'04)\/}, volume~2, pages 97--104, Los Alamitos, CA, USA. IEEE
  Computer Society.

\bibitem[Netzer {\em et~al.}(2011)Netzer, Wang, Coates, Bissacco, Wu, and
  Ng]{Netzer-wkshp-2011}
Netzer, Y., Wang, T., Coates, A., Bissacco, A., Wu, B., and Ng, A.~Y. (2011).
\newblock Reading digits in natural images with unsupervised feature learning.
\newblock Deep Learning and Unsupervised Feature Learning Workshop, {NIPS}.

\bibitem[Oliphant(2007)Oliphant]{numpy-2007}
Oliphant, T.~E. (2007).
\newblock Python for scientific computing.
\newblock {\em Computing in Science and Engineering\/}, {\bf 9}, 10--20.

\bibitem[Pearson(1901)Pearson]{Pearson-1901}
Pearson, K. (1901).
\newblock On lines and planes of closest fit to systems of points in space.
\newblock {\em Philosophical Magazine\/}, {\bf 2}(6), 559--572.

\bibitem[Pedregosa {\em et~al.}(2011)Pedregosa, Varoquaux, Gramfort, Michel,
  Thirion, Grisel, Blondel, Prettenhofer, Weiss, Dubourg, Vanderplas, Passos,
  Cournapeau, Brucher, Perrot, and Duchesnay]{scikit-learn}
Pedregosa, F., Varoquaux, G., Gramfort, A., Michel, V., Thirion, B., Grisel,
  O., Blondel, M., Prettenhofer, P., Weiss, R., Dubourg, V., Vanderplas, J.,
  Passos, A., Cournapeau, D., Brucher, M., Perrot, M., and Duchesnay, E.
  (2011).
\newblock Scikit-learn: Machine learning in {P}ython.
\newblock {\em Journal of Machine Learning Research\/}, {\bf 12}, 2825--2830.

\bibitem[Polak and Ribiere(1969)Polak and Ribiere]{polak1969note}
Polak, E. and Ribiere, G. (1969).
\newblock {Note sur la convergence de m{\'{e}}thodes de directions
  conjugu{\'{e}}es}.
\newblock {\em Revue Fran{\c{c}}aise d'Informatique et de Recherche
  Op{\'{e}}rationnelle\/}, {\bf 16}, 35--43.

\bibitem[Polyak and Juditsky(1992)Polyak and Juditsky]{Polyak+Juditsky-1992}
Polyak, B. and Juditsky, A. (1992).
\newblock Acceleration of stochastic approximation by averaging.
\newblock {\em {SIAM} J. Control and Optimization\/}, {\bf 30(4)}, 838--855.

\bibitem[Rifai {\em et~al.}(2011a)Rifai, Vincent, Muller, Glorot, and
  Bengio]{Rifai+al-2011}
Rifai, S., Vincent, P., Muller, X., Glorot, X., and Bengio, Y. (2011a).
\newblock Contractive auto-encoders: Explicit invariance during feature
  extraction.
\newblock In {\em Proceedings of theTwenty-eight International Conference on
  Machine Learning (ICML'11)\/}.

\bibitem[Rifai {\em et~al.}(2011b)Rifai, Mesnil, Vincent, Muller, Bengio,
  Dauphin, and Glorot]{Salah+al-2011}
Rifai, S., Mesnil, G., Vincent, P., Muller, X., Bengio, Y., Dauphin, Y., and
  Glorot, X. (2011b).
\newblock Higher order contractive auto-encoder.
\newblock In {\em European Conference on Machine Learning and Principles and
  Practice of Knowledge Discovery in Databases (ECML PKDD)\/}.

\bibitem[Rumelhart {\em et~al.}(1986)Rumelhart, Hinton, and
  Williams]{Rumelhart86c}
Rumelhart, D.~E., Hinton, G.~E., and Williams, R.~J. (1986).
\newblock Learning internal representations by error propagation.
\newblock In D.~E. Rumelhart and J.~L. McClelland, editors, {\em Parallel
  Distributed Processing\/}, volume~1, chapter~8, pages 318--362. MIT Press,
  Cambridge.

\bibitem[Salakhutdinov and Hinton(2009)Salakhutdinov and
  Hinton]{SalHinton09small}
Salakhutdinov, R. and Hinton, G. (2009).
\newblock Deep {B}oltzmann machines.
\newblock In {\em AISTATS'2009\/}, pages 448--455.

\bibitem[Sermanet {\em et~al.}(2012)Sermanet, Chintala, and
  LeCun]{sermanet-icpr-12}
Sermanet, P., Chintala, S., and LeCun, Y. (2012).
\newblock Convolutional neural networks applied to house numbers digit
  classification.
\newblock In {\em International Conference on Pattern Recognition (ICPR
  2012)\/}.

\bibitem[Smolensky(1986)Smolensky]{Smolensky86}
Smolensky, P. (1986).
\newblock Information processing in dynamical systems: Foundations of harmony
  theory.
\newblock In D.~E. Rumelhart and J.~L. McClelland, editors, {\em Parallel
  Distributed Processing\/}, volume~1, chapter~6, pages 194--281. MIT Press,
  Cambridge.

\bibitem[Srebro and Shraibman(2005)Srebro and Shraibman]{Srebro05}
Srebro, N. and Shraibman, A. (2005).
\newblock Rank, trace-norm and max-norm.
\newblock In {\em Proceedings of the 18th Annual Conference on Learning
  Theory\/}, pages 545--560. Springer-Verlag.

\bibitem[Steinhaus(1957)Steinhaus]{Steinhaus57}
Steinhaus, H. (1957).
\newblock Sur la division des corps mat\'eriels en parties.
\newblock In {\em Bull. Acad. Polon. Sci.}, pages 801--804.

\bibitem[Susskind {\em et~al.}(2010)Susskind, Anderson, and
  Hinton]{Susskind2010}
Susskind, J., Anderson, A., and Hinton, G.~E. (2010).
\newblock The {T}oronto face dataset.
\newblock Technical Report UTML TR 2010-001, U. Toronto.

\bibitem[Tieleman(2008)Tieleman]{Tieleman08}
Tieleman, T. (2008).
\newblock Training restricted {B}oltzmann machines using approximations to the
  likelihood gradient.
\newblock In W.~W. Cohen, A.~McCallum, and S.~T. Roweis, editors, {\em {ICML}
  2008\/}, pages 1064--1071. ACM.

\bibitem[Vincent(2011)Vincent]{Vincent-NC-2011}
Vincent, P. (2011).
\newblock A connection between score matching and denoising autoencoders.
\newblock {\em Neural Computation\/}, {\bf 23}(7), 1661--1674.

\bibitem[Vincent {\em et~al.}(2008)Vincent, Larochelle, Bengio, and
  Manzagol]{VincentPLarochelleH2008}
Vincent, P., Larochelle, H., Bengio, Y., and Manzagol, P.-A. (2008).
\newblock Extracting and composing robust features with denoising autoencoders.
\newblock In {\em ICML'08\/}, pages 1096--1103. ACM.

\bibitem[Welling {\em et~al.}(2005)Welling, Rosen-Zvi, and Hinton]{Welling05}
Welling, M., Rosen-Zvi, M., and Hinton, G.~E. (2005).
\newblock Exponential family harmoniums with an application to information
  retrieval.
\newblock In {\em NIPS'04\/}, volume~17, Cambridge, MA. MIT Press.

\bibitem[Younes(1999)Younes]{Younes1999}
Younes, L. (1999).
\newblock On the convergence of {M}arkovian stochastic algorithms with rapidly
  decreasing ergodicity rates.
\newblock {\em Stochastics and Stochastic Reports\/}, {\bf 65}(3), 177--228.

\bibitem[Yu {\em et~al.}(2009)Yu, Zhang, and Gong]{Yu+al-NIPS09}
Yu, K., Zhang, T., and Gong, Y. (2009).
\newblock Nonlinear learning using local coordinate coding.
\newblock In Y.~Bengio, D.~Schuurmans, J.~Lafferty, C.~K.~I. Williams, and
  A.~Culotta, editors, {\em Advances in Neural Information Processing Systems
  22\/}, pages 2223--2231.

\end{thebibliography}
\bibliographystyle{natbib}

\end{document}